# Word-Sense Disambiguation Using Decomposable Models

**Rebecca Bruce** and **Janyce Wiebe**
Computing Research Lab
and
Department of Computer Science
New Mexico State University
Las Cruces, NM 88003
rbruce@cs.nmsu.edu, wiebe@cs.nmsu.edu

## Abstract

Most probabilistic classifiers used for word-sense disambiguation have either been based on only one contextual feature or have used a model that is simply assumed to characterize the interdependencies among multiple contextual features. In this paper, a different approach to formulating a probabilistic model is presented along with a case study of the performance of models produced in this manner for the disambiguation of the noun *interest*. We describe a method for formulating probabilistic models that use multiple contextual features for word-sense disambiguation, without requiring untested assumptions regarding the form of the model. Using this approach, the joint distribution of all variables is described by only the most systematic variable interactions, thereby limiting the number of parameters to be estimated, supporting computational efficiency, and providing an understanding of the data.

## Introduction

This paper presents a method for constructing probabilistic classifiers for word-sense disambiguation that offers advantages over previous approaches. Most previous efforts have not attempted to systematically identify the interdependencies among contextual features (such as collocations) that can be used to classify the meaning of an ambiguous word. Many researchers have performed disambiguation on the basis of only a single feature, while others who do consider multiple contextual features assume that all contextual features are either conditionally independent given the sense of the word or fully independent. Of course, all contextual features could be treated as interdependent, but, if there are several features, such a model could have too many parameters to estimate in practice.

We present a method for formulating probabilistic models that describe the relationships among all variables in terms of only the most important interdependencies, that is, models of a certain class that are good approximations to the joint distribution of contextual features and word meanings. This class is the set of *decomposable models*: models that can be expressed as a product of marginal distributions, where each marginal is composed of interdependent variables. The test used to evaluate a model gives preference to those that have the fewest number of interdependencies, thereby selecting models expressing only the most systematic variable interactions.

To summarize the method, one first identifies informative contextual features (where "informative" is a well-defined notion, discussed in Section 2). Then, out of all possible decomposable models characterizing interdependency relationships among the selected variables, those that are found to produce good approximations to the data are identified (using the test mentioned above) and one of those models is used to perform disambiguation. Thus, we are able to use multiple contextual features without the need for untested assumptions regarding the form of the model. Further, approximating the joint distribution of all variables with a model identifying only the most important systematic interactions among variables limits the number of parameters to be estimated, supports computational efficiency, and provides an understanding of the data. The biggest limitation associated with this method is the need for large amounts of sense-tagged data. Because asymptotic distributions of the test statistics are used, the validity of the results obtained using this approach are compromised when it is applied to sparse data (this point is discussed further in Section 2).

To test the method of model selection presented in this paper, a case study of the disambiguation of the noun *interest* was performed. *Interest* was selected because it has been shown in previous studies to be a difficult word to disambiguate. We selected as the set of sense tags all non-idiomatic noun senses of *interest* defined in the electronic version of Longman's Dictionary of Contemporary English (LDOCE) ([23]). Using the models produced in this study, we are able to assign an LDOCE sense tag to every usage of *interest* in a held-out test set with 78% accuracy. Although it is difficult to compare our results to those reported for previous disambiguation experiments, as will be discussed later, we feel these results are encouraging.

The remainder of the paper is organized as follows. Section 2 provides a more complete definition of the methodology used for formulating decomposable models and Section 3 describes the details of the case study performed to test the approach. The results of the disambiguation case study are discussed and contrasted with similar efforts in Sections 4 and 5. Section 6 is the conclusion.

## Decomposable Models

In this Section, we address the problem of finding the models that generate good approximations to a given discrete probability distribution, as selected from among the class of *decomposable models*. Decomposable models are a subclass of log-linear models and, as such, can be used to characterize and study the structure of data ([2]), that is, the interactions among variables as evidenced by the frequency with which the values of the variables co-occur. Given a data sample of objects, where each object is described by $d$ discrete variables, let $\mathbf{x}=(x_1, x_2, \ldots, x_q)$ be a $q$-dimensional vector of counts, where each $x_i$ is the frequency with which one of the possible combinations of the values of the $d$ variables occurs in the data sample (and the frequencies of all such possible combinations are included in $\mathbf{x}$). The log-linear model expresses the logarithm of $E[\mathbf{x}]$ (the mean of $\mathbf{x}$) as a linear sum of the contributions of the "effects" of the variables and the interactions among the variables.

Assume that a random sample consisting of $N$ independent and identical trials (i.e., all trials are described by the same probability density function) is drawn from a discrete $d$-variate distribution. In such a situation, the outcome of each trial must be an event corresponding to a particular combination of the values of the $d$ variables. Let $p_i$ be the probability that the $i^{th}$ event (i.e., the $i^{th}$ possible combination of the values of all variables) occurs on any trial and let $x_i$ be the number of times that the $i^{th}$ event occurs in the random sample. Then $(x_1, x_2, \ldots, x_q)$ has a multinomial distribution with parameters $N$ and $p_1, \ldots, p_q$. For a given sample size, $N$, the likelihood of selecting any particular random sample is defined once the population parameters, that is, the $p_i$'s or, equivalently, the $E[x_i]$'s (where $E[x_i]$ is the mean frequency of event $i$), are known. Log-linear models express the value of the logarithm of each $E[x_i]$ or $p_i$ as a linear sum of a smaller (i.e., less than $q$) number of new population parameters that characterize the effects of individual variables and their interactions.

The theory of log-linear models specifies the *sufficient statistics* (functions of $\mathbf{x}$) for estimating the effects of each variable and of each interaction among variables on $E[\mathbf{x}]$. The sufficient statistics are the sample counts from the highest-order marginals composed of only interdependent variables. These statistics are the maximum likelihood estimates of the mean values of the corresponding marginals distributions. Consider, for example, a random sample taken from a population in which four contextual features are used to characterize each occurrence of an ambiguous word. The sufficient statistics for the model describing contextual features one and two as independent but all other variables as interdependent are, for all $i, j, k, m, n$ (in this and all subsequent equations, $f$ is an abbreviation for *feature*):

$$\hat{E}[\text{count}(f_2 = j, f_3 = k, f_4 = m, tag = n)] =$$
$$\sum_i x_{f_1=i, f_2=j, f_3=k, f_4=m, tag=n}$$

and

$$\hat{E}[\text{count}(f_1 = i, f_3 = k, f_4 = m, tag = n)] =$$
$$\sum_j x_{f_1=i, f_2=j, f_3=k, f_4=m, tag=n}$$

Within the class of decomposable models, the maximum likelihood estimate for $E[\mathbf{x}]$ reduces to the product of the sufficient statistics divided by the sample counts defined in the marginals composed of the common elements in the sufficient statistics. As such, decomposable models are models that can be expressed as a product of marginals,[1] where each marginal consists of only interdependent variables.

Returning to our previous example, the maximum likelihood estimate for $E[\mathbf{x}]$ is, for all $i, j, k, m, n$:

$$\hat{E}[x_{f_1=i, f_2=j, f_3=k, f_4=m, tag=n}] =$$
$$\quad \hat{E}[\text{count}(f_1 = i, f_3 = k, f_4 = m, tag = n)] \times$$
$$\quad \hat{E}[\text{count}(f_2 = j, f_3 = k, f_4 = m, tag = n)] \div$$
$$\quad \hat{E}[\text{count}(f_3 = k, f_4 = m, tag = n)]$$

Expressing the population parameters as probabilities instead of expected counts, the equation above can be rewritten as follows, where the sample marginal relative frequencies are the maximum likelihood estimates of the population marginal probabilities. For all $i, j, k, m, n$:

$$\hat{P}(f_1 = i, f_2 = j, f_3 = k, f_4 = m, tag = n) =$$
$$\quad \hat{P}(f_1 = i \mid f_3 = k, f_4 = m, tag = n) \times$$
$$\quad \hat{P}(f_2 = j \mid f_3 = k, f_4 = m, tag = n) \times$$
$$\quad \hat{P}(f_3 = k, f_4 = m, tag = n)$$

The degree to which the data is approximated by a model is called the *fit* of the model. In this work, the likelihood ratio statistic, $G^2$, is used as the measure of the goodness-of-fit of a model. It is distributed asymptotically as $\chi^2$ with degrees of freedom corresponding to the number of interactions (and/or variables) omitted from (unconstrained in) the model. Accessing the fit of a model in terms of the significance of its $G^2$ statistic gives preference to models with the fewest number of interdependencies, thereby assuring the selection of a model specifying only the most systematic variable interactions.

---

[1] The marginal distributions can be represented in terms of counts or relative frequencies, depending on whether the parameters are expressed as expected frequencies or probabilities, respectively.

Within the framework described above, the process of model selection becomes one of hypothesis testing, where each pattern of dependencies among variables expressible in terms of a decomposable model is postulated as a hypothetical model and its fit to the data is evaluated. The "best fitting" models are identified, in the sense that the significance of their reference $\chi^2$ values are large, and, from among this set, a conceptually appealing model is chosen. The exhaustive search of decomposable models can be conducted as described in [12].

What we have just described is a method for approximating the joint distribution of all variables with a model containing only the most important systematic interactions among variables. This approach to model formulation limits the number of parameters to be estimated, supports computational efficiency, and provides an understanding of the data. The single biggest limitation remaining in this day of large memory, high speed computers results from reliance on asymptotic theory to describe the distribution of the maximum likelihood estimates and the likelihood ratio statistic. The effect of this reliance is felt most acutely when working with large sparse multinomials, which is exactly when this approach to model construction is most needed. When the data is sparse, the usual asymptotic properties of the distribution of the likelihood ratio statistic and the maximum likelihood estimates may not hold. In such cases, the fit of the model will appear to be too good, indicating that the model is in fact over constrained for the data available. In this work, we have limited ourselves to considering only those models with sufficient statistics that are not sparse, where the significance of the reference $\chi^2$ is not unreasonable; most such models have sufficient statistics that are lower-order marginal distributions. In the future, we will investigate other goodness-of-fit tests ([18], [1], [22]) that are perhaps more appropriate for sparse data.

## The Experiment

Unlike several previous approaches to word sense disambiguation ([29], [5], [7], [10]), nothing in this approach limits the selection of sense tags to a particular number or type of meaning distinctions. In this study, our goal was to address a non-trivial case of ambiguity, but one that would allow some comparison of results with previous work. As a result of these considerations, the word *interest* was chosen as a test case, and the six non-idiomatic noun senses of *interest* defined in LDOCE were selected as the tag set. The only restriction limiting the choice of corpus is the need for large amounts of on-line data. Due to availability, the Penn Treebank Wall Street Journal corpus was selected.

In total, 2,476 usages[2] of *interest* as a noun[3] were automatically extracted from the corpus and manually assigned sense tags corresponding to the LDOCE definitions.

During tagging, 107 usages were removed from the data set due to the authors' inability to classify them in terms of the set of LDOCE senses. Of the rejected usages, 43 are metonymic, and the rest are hybrid meanings specific to the domain, such as *public interest group*.

Because our sense distinctions are not merely between two or three clearly defined core senses of a word, the task of hand-tagging the tokens of *interest* required subtle judgments, a point that has also been observed by other researchers disambiguating with respect to the full set of LDOCE senses ([6], [28]). Although this undoubtedly degraded the accuracy of the manually assigned sense tags (and thus the accuracy of the study as well), this problem seems unavoidable when making semantic distinctions beyond clearly defined core senses of a word ([17], [11], [14], [15]).

Of the 2,369 sentences containing the sense-tagged usages of *interest*, 600 were randomly selected and set aside to serve as the test set. The distribution of sense tags in the data set is presented in Table 1.

We now turn to the selection of individually informative contextual features. In our approach to disambiguation, a contextual feature is judged to be informative (i.e., correlated with the sense tag of the ambiguous word) if the model for independence between that feature and the sense tag is judged to have an extremely poor fit using the test described in Section 2. The worse the fit, the more informative the feature is judged to be (similar to the approach suggested in [9]).

Only features whose values can be automatically determined were considered, and preference was given to features that intuitively are not specific to *interest* (but see the discussion of collocational features below). An additional criterion was that the features not have too many possible values, in order to curtail sparsity in the resulting data matrix.

We considered three different types of contextual features: morphological, collocation-specific, and class-based, with part-of-speech (POS) categories serving as the word classes. Within these classes, we choose a number of specific features, each of which was judged to be informative as described above. We used one morphological feature: a dichotomous variable indicating the presence or absence of the plural form. The values of the class-based variables are a set of twenty-five POS tags formed, with one exception, from the first letter of the tags used in the Penn Treebank corpus. Two different sets of class-based variables were selected. The first set contained only the POS tags of the word immediately preceding and the word immediately succeeding the ambiguous word, while the second set was extended to include the POS tags of the two immediately preceding and two succeeding words.

---

[2] For sentences with more than one usage, the tool used to automatically extract the test data ignored all but one of them. Thus, some usages were missed.

[3] The Penn Treebank corpus comes complete with POS tags.

A limited number of collocation-specific variables were selected, where the term *collocation* is used loosely to refer to a specific spelling form occurring in the same sentence as the ambiguous word. All of our collocational variables are dichotomous, indicating the presence or absence of the associated spelling form. While collocation-specific variables are, by definition, specific to the word being disambiguated, the procedure used to select them is general. The search for collocation-specific variables was limited to the 400 most frequent spelling forms in a data sample composed of sentences containing *interest*. Out of these 400, the five spelling forms found to be the most informative using the test described above were selected as the collocational variables.

It is not enough to know that each of the features described above is highly correlated with the meaning of the ambiguous word. In order to use the features in concert to perform disambiguation, a model describing the interactions *among* them is needed. Since we had no reason to prefer, a priori, one form of model over another, all models describing possible interactions among the features were generated, and a model with good fit was selected. Models were generated and tested as described in Section 2.

## Results

Both the form and the performance of the model selected for each set of variables is presented in Table 2. Performance is measured in terms of the total percentage of the test set tagged correctly by a classifier using the specified model. This measure combines both precision and recall. Portions of the test set that are not covered by the estimates of the parameters made from the training set are not tagged and, therefore, counted as wrong.

The form of the model describes the interactions among the variables by expressing the joint distribution of the values of all contextual features and sense tags as a product of conditionally independent marginals, with each marginal being composed of non-independent variables. Models of this form describe a markov field ([8], [21]) that can be represented graphically as is shown in Figure 1 for Model 4 of Table 2. In both Figures 1 and 2, each of the variables *short, in, pursue, rate(s), percent* (i.e., the sign '%') is the presence or absence of that spelling form. Each of the variables *r1pos, r2pos, l1pos*, and *l2pos* is the POS tag of the word 1 or 2 positions to the left (*l*) or right (*r*). The variable *ending* is whether *interest* is in the singular or plural, and the variable *tag* is the sense tag assigned to *interest*.

The graphical representation of Model 4 is such that there is a one-to-one correspondence between the nodes of the graph and the sets of conditionally independent variables in the model. The semantics of the graph topology is that all variables that are not directly connected in the graph are conditionally independent given the values of the variables mapping to the connecting nodes. For example, if node $a$ separates node $b$ from node $c$ in the graphical representation of a markov field, then the variables mapping to node $b$ are conditionally independent of the variables mapping to node $c$ given the values of the variables mapping to node $a$. In the case of Model 4, Figure 1 graphically depicts the fact that the value of the morphological variable *ending* is conditionally independent of the values of all other contextual features given the sense tag of the ambiguous word.

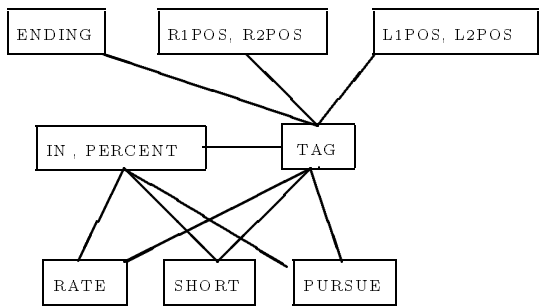

Figure 1

The Markov field depicted in Figure 1 is represented by an undirected graph because conditional independence is a symmetric relationship. But decomposable models can also be characterized by directed graphs and interpreted according to the semantics of a Bayesian network ([21]; also described as "recursive causal models" in [27] and [16]). In a Bayesian network, the notions of causation and influence replace the notion of conditional independence in a Markov field. The parents of a variable (or set of variables) $V$ are those variables judged to be the *direct causes* or to have *direct influence* on the value of $V$; $V$ is called a "response" to those causes or influences. The Bayesian network representation of a decomposable model embodies an explicit ordering of the $n$ variables in the model such that variable $i$ may be considered a response to some or all of variables $\{i+1, \ldots, n\}$, but is not thought of as a response to any one of the variables $\{1, \ldots, i-1\}$. In all models presented in this paper, the sense tag of the ambiguous word causes or influences the values of all other variables in the model. The Bayesian network representation of Model 4 is presented in Figure 2. In Model 4, the variables *in* and *percent* are treated as influencing the values of *rate, short,* and *pursue* in order to achieve an ordering of variables as described above.

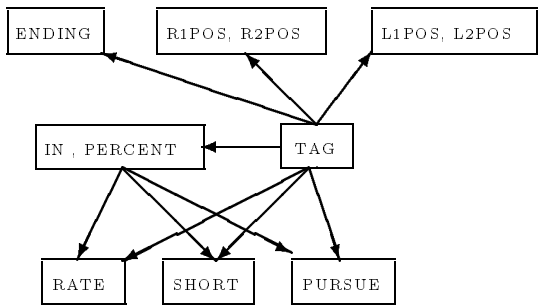

Figure 2

## Comparison to Previous Work

Many researchers have avoided characterizing the interactions among multiple contextual features by considering only one feature in determining the sense of an ambiguous word. Techniques for identifying the optimum feature to use in disambiguating a word are presented in [7], [30] and [5]. Other works consider multiple contextual features in performing disambiguation without formally characterizing the relationships among the features. The majority of these efforts ([13], [31]) weight each feature in predicting the sense of an ambiguous word in accordance with frequency information, without considering the extent to which the features co-occur with one another. Gale, Church and Yarowsky ([10]) and Yarowsky ([29]) formally characterize the interactions that they consider in their model, but they simply *assume* that their model fits the data.

Other researchers have proposed approaches to systematically combining information from multiple contextual features in determining the sense of an ambiguous word. Schutze ([26]) derived contextual features from a singular value decomposition of a matrix of letter four-gram co-occurrence frequencies, thereby assuring the independence of all features. Unfortunately, interpreting a contextual feature that is a weighted combination of letter four-grams is difficult. Further, the clustering procedure used to assign word meaning based on these features is such that the resulting sense clusters do not have known statistical properties. This makes it impossible to generalize the results to other data sets.

Black ([3]) used decision trees ([4]) to define the relationships among a number of pre-specified contextual features, which he called "contextual categories", and the sense tags of an ambiguous word. The tree construction process used by Black partitions the data according to the values of one contextual feature before considering the values of the next, thereby treating all features incorporated in the tree as interdependent. The method presented here for using information from multiple contextual features is more flexible and makes better use of a small data set by eliminating the need to treat all features as interdependent.

The work that bears the closest resemblance to the work presented here is the maximum entropy approach to developing language models ([24], [25], [19] and [20]). Although this approach has not been applied to word-sense disambiguation, there is a strong similarity between that method of model formulation and our own. A maximum entropy model for multivariate data is the likelihood function with the highest entropy that satisfies a pre-defined set of linear constraints on the underlying probability estimates. The constraints describe interactions among variables by specifying the expected frequency with which the values of the constrained variables co-occur. When the expected frequencies specified in the constraints are linear combinations of the observed frequencies in the training data, the resulting maximum entropy model is equivalent to a maximum likelihood model, which is the type of model used here.

To date, in the area of natural language processing, the principles underlying the formulation of maximum entropy models have been used only to estimate the parameters of a model. Although the method described in this paper for finding a good approximation to the joint distribution of a set of discrete variables makes use of maximum likelihood models, the scope of the technique we are describing extends beyond parameter estimation to include selecting the form of the model that approximates the joint distribution.

Several of the studies mentioned in this Section have used *interest* as a test case, and all of them (with the exception of Schutze [26]) considered four possible meanings for that word. In order to facilitate comparison of our work with previous studies, we re-estimated the parameters of our best model and tested it using data containing only the four LDOCE senses corresponding to those used by others (usages not tagged as being one of these four senses were removed from both the test and training data sets). The results of the modified experiment along with a summary of the published results of previous studies are presented in Table 3.

While it is true that all of the studies reported in Table 3 used four senses of *interest*, it is not clear that any of the other experimental parameters were held constant in all studies. Therefore, this comparison is only suggestive. In order to facilitate more meaningful comparisons in the future, we are donating the data used in this experiment to the Consortium for Lexical Research (ftp site: clr.nmsu.edu) where it will be available to all interested parties.

## Conclusions and Future Work

In this paper, we presented a method for formulating probabilistic models that use multiple contextual features for word-sense disambiguation without requiring untested assumptions regarding the form of the model. In this approach, the joint distribution of all variables is described by only the most systematic variable interactions, thereby limiting the number of parameters to be estimated, supporting computational efficiency, and providing an understanding of the data. Further, different types of variables, such as class-based and collocation-specific ones, can be used in combination with one another. We also presented the results of a study testing this approach. The results suggest that the models produced in this study perform as well as or better than previous efforts on a difficult test case.

We are investigating several extensions to this work. In order to reasonably consider doing large-scale word-sense disambiguation, it is necessary to eliminate the need for large amounts of manually sense-tagged data. In the future, we hope to develop a parametric model or models applicable to a wide range of content words and to estimate the parameters of those models from untagged data. To those ends, we are currently investigating a means of obtaining maximum likelihood estimates of the parameters of decomposable models from untagged data. The procedure we are using is a variant of the EM algorithm that is specific to models of the form produced in this study. Preliminary results are mixed, with performance being reasonably good on models with low-order marginals (e.g., 63% of the test set was tagged correctly with Model 1 using parameters estimated in this manner) but poorer on models with higher-order marginals, such as Model 4. Work is needed to identify and constrain the parameters that cannot be estimated from the available data and to determine the amount of data needed for this procedure.

We also hope to integrate probabilistic disambiguation models, of the type described in this paper, with a constraint-based knowledge base such as WordNet. In the past, there have been two types of approaches to word sense disambiguation: 1) a probabilistic approach such as that described here which bases the choice of sense tag on the observed joint distribution of the tags and contextual features, and 2) a symbolic knowledge based approach that postulates some kind of relational or constraint structure among the words to be tagged. We hope to combine these methodologies and thereby derive the benefits of both. Our approach to combining these two paradigms hinges on the network representations of our probabilistic models as described in Section 4 and will make use of the methods presented in [21].

## Acknowledgements

The authors would like to thank Gerald Rogers for sharing his expertise in statistics, Ted Dunning for advice and support on software development, and the members of the NLP group in the CRL for helpful discussions.

## References

[1] Baglivo, J., Olivier, D., and Pagano, M. (1992). Methods for Exact Goodness-of-Fit Tests. *Journal of the American Statistical Association*, Vol. 87, No. 418, June 1992.

[2] Bishop, Y. M.; Fienberg, S.; and Holland, P (1975). *Discrete Multivariate Analysis: Theory and Practice.* Cambridge: The MIT Press.

[3] Black, Ezra (1988). An Experiment in Computational Discrimination of English Word Senses. *IBM Journal of Research and Development*, Vol. 32, No. 2, pp. 185-194.

[4] Breiman, L., Friedman, J., Olshen, R., and Stone, C. (1984). *Classification and Regression Trees.* Monterey, CA: Wadsworth & Brooks/Cole Advanced Books & Software.

[5] Brown, P., Della Pietra, S., Della Pietra, V., and Mercer, R. (1991). Word Sense Disambiguation Using Statistical Methods. *Proceedings of the 29th Annual Meeting of the Association for Computational Linguistics (ACL-91)*, pp. 264-304.

[6] Cowie, J., Guthrie, J., and Guthrie, L. (1992). Lexical Disambiguation and Simulating Annealing. *Proceedings of the 15th International Conference on Computational Linguistics (COLING-92)*. pp 359-365.

[7] Dagan, I., Itai, A., and Schwall, U. (1991). Two Languages Are More Informative Than One. *Proceedings of the 29th Annual Meeting of the Association for Computational Linguistics (ACL-91)*, pp. 130-137.

[8] Darroch, J., Lauritzen, S., and Speed, T. (1980). Markov Fields and Log-Linear Interaction Models for Contingency Tables. *The Annals of Statistics*, Vol. 8, No. 3, pp. 522-539.

[9] Dunning, Ted (1993). Accurate Methods for the Statistics of Surprise and Coincidence. *Computational Linguistics*, Vol. 19, No. 1, pp.61-74.

[10] Gale, W., Church, K., and Yarowsky, D. (1992a). A Method for Disambiguating Word Senses in a Large Corpus. *AT&T Bell Laboratories Statistical Research Report No. 104*.

[11] Gale, W., Church, K. and Yarowsky, D. (1992b). Estimating Upper and Lower Bounds on the Performance of Word-Sense Disambiguation Programs. *Proceedings of the 30th Annual Meeting of the ACL*, 1992.

[12] Havranek, Tomas (1984). A Procedure for Model Search in Multidimensional Contingency Tables. *Biometrics*, 40, pp.95-100.

[13] Hearst, Marti (1991). Toward Noun Homonym Disambiguation—Using Local Context in Large Text Corpora. *Proceedings of the Seventh Annual Conference of the UW Centre for the New OED and Text Research Using Corpora*, pp. 1-22.

[14] Jorgensen, Julia (1990). The Psychological Reality of Word Senses. *Journal of Psycholinguistic Research*, Vol 19, pp. 167-190.

[15] Kelly, E and P. Stone (1979). *Computer Recognition of English Word Senses*, Vol. 3 of *North Holland Linguistics Series*, Amsterdam: North-Holland.

[16] Kiiveri, H., Speed, T., and Carlin, J. (1984). Recursive Causal Models. *Journal Austral. Math. Soc.* (Series A), 36, pp. 30-52.

[17] Kilgarriff, Adam (1993). Dictionary Word Sense Distinctions: An Enquiry Into Their Nature. *Computers and the Humanities*, 26, pp.365-387.

[18] Koehler, K. (1986). Goodness-of-Fit Tests for Log-Linear Models in Sparse Contingency Tables. *Journal of the American Statistical Association*, Vol. 81, No. 394, June 1986.

[19] Lau, R., Rosenfeld, R., and Roukos, S. (1993a). Trigger-Based Language Models: a Maximum Entropy Approach. *Proceedings of ICASSP-93*. April 1993.

[20] Lau, R., Rosenfeld, R., and Roukos, S. (1993b). Adaptive Language Modeling Using the Maximum Entropy Principle. *Proc. ARPA Human Language Technology Workshop*. March 1993.

[21] Pearl, Judea (1988). *Probabilistic Reasoning In Intelligent Systems: Networks of Plausible Inference*. San Mateo, Ca.: Morgan Kaufmann.

[22] Pederson, S. and Johnson, M. (1990). Estimating Model Discrepancy. *Technometrics*, Vol. 32, No. 3, pp. 305-314.

[23] Procter, Paul et al. (1978). *Longman Dictionary of Contemporary English*.

[24] Ratnaparkhi, A. and Roukos, S. (1994). A Maximum Entropy Model for Prepositional Phrase Attachment. *Proc. ARPA Human Language Technology Workshop*. March 1994.

[25] Rosenfeld, R. (1994). A Hybrid Approach to Adaptive Statistical Language Modeling. *Proc. ARPA Human Language Technology Workshop*. March 1994.

[26] Schutze, Hinrich (1992). Word Space. In S.J. Hanson, J.D. Cowan, and C.L. Giles (Eds.), *Advances in Neural Information Processing Systems 5*, San Mateo, Ca.: Morgan Kaufmann.

[27] Wermuth, N. and Lauritzen, S. (1983). Graphical and recursive models for contingency tables. *Biometrika*, Vol. 70, No. 3, pp. 537-52.

[28] Wilks, Y., Fass, D., Guo, C., McDonald, J., Plate, T., and Slator, B. (1990). Providing Machine Tractable Dictionary Tools. *Computers and Translation 2*. Also to appear in *Theoretical and Computational Issues in Lexical Semantics (TCILS)*. Edited by James Pustejovsky. Cambridge, MA.: MIT Press.

[29] Yarowsky, David (1992). Word-Sense Disambiguating Using Statistical Models of Roget's Categories Trained on Large Corpora. *Proceedings of the 15th International Conference on Computational Linguistics (COLING-92)*.

[30] Yarowsky, David (1993). One Sense Per Collocation. *Proceedings of the Speech and Natural Language ARPA Workshop*, March 1993, Princeton, NJ.

[31] Zernik, Uri (1990). Tagging Word Senses In Corpus: The Needle in the Haystack Revisited. *Technical Report 90CRD198*, GE Research and Development Center.

| LDOCE Sense | Representation in Total Sample | Representation in Training Sample | Representation in Test Sample |
|---|---|---|---|
| sense 1: "readiness to give attention" | 361 (15%) | 271 (15%) | 90 (15%) |
| sense 2: "quality of causing attention to be given" | 11 (<1%) | 9 (<1%) | 2 (<1%) |
| sense 3: "activity, subject, etc., which one gives time and attention to" | 66 (3%) | 50 (3%) | 16 (3%) |
| sense 4: "advantage, advancement, or favor" | 178 (8%) | 130 (7%) | 48 (8%) |
| sense 5: "a share (in a company, business, etc.)" | 500 (21%) | 378 (21%) | 122 (20%) |
| sense 6: "money paid for the use of money" | 1253 (53%) | 931 (53%) | 322 (54%) |

Table 1: Distribution of sense tags.

| | Model | Percent Correct |
|---|---|---|
| 1 | $P(r1pos, l1pos, ending, tag) =$ $P(r1pos|tag) \times P(l1pos|tag) \times P(ending|tag) \times P(tag)$ | 73% |
| 2 | $P(r1pos, r2pos, l1pos, l2pos, ending, tag) =$ $P(r1pos, r2pos|tag) \times P(l1pos, l2pos|tag) \times P(ending|tag) \times P(tag)$ | 76% |
| 3 | $P(percent, pursue, short, in, rate, tag) =$ $P(short|percent, in, tag) \times P(rate|percent, in, tag) \times$ $P(pursue|percent, in, tag) \times P(percent, in|tag) \times P(tag)$ | 61% |
| 4 | $P(percent, pursue, short, in, rate, r1pos, r2pos, l1pos, l2pos, ending, tag) =$ $P(short|percent, in, tag) \times P(rate|percent, in, tag) \times P(pursue|percent, in, tag) \times$ $P(percent, in|tag) \times P(r1pos, r2pos|tag) \times P(l1pos, l2pos|tag) \times P(ending|tag) \times P(tag)$ | 78% |

Table 2: The form and performance on the test data of the model found for each set of variables. Each of the variables *short*, *in*, *pursue*, *rate(s)*, *percent* (i.e., the sign '%') is the presence or absence of that spelling form. Each of the variables *r1pos*, *r2pos*, *l1pos*, and *l2pos* is the POS tag of the word 1 or 2 positions to the left (*l*) or right (*r*). The variable *ending* is whether *interest* is in the singular or plural, and the variable *tag* is the sense tag assigned to *interest*.

| Model | Percent Correct |
|---|---|
| Black (1988) | 72% |
| Zernik (1990) | 70% |
| Yarowsky (1992) | 72% |
| Bruce & Wiebe model 4 using only four senses | 79% |

Table 3: Comparison to previous results.